\documentclass[twoside,leqno,twocolumn]{article}

\usepackage[letterpaper]{geometry}

\usepackage{ltexpprt}
\usepackage{hyperref}

\usepackage{times}
\usepackage{amsfonts}
\usepackage{url}
\usepackage[utf8]{inputenc}
\usepackage{caption}
\usepackage{natbib}
\usepackage{graphicx}
\usepackage{amsmath}

\newtheorem{definition}{Definition}
\usepackage{array,multirow}
\usepackage{multicol}
\usepackage{booktabs}
\usepackage{subcaption}
\usepackage{xcolor}

\usepackage{dsfont}
\newcolumntype{P}[1]{>{\centering\arraybackslash}p{#1}}
\def\BibTeX{{\rm B\kern-.05em{\sc i\kern-.025em b}\kern-.08em
    T\kern-.1667em\lower.7ex\hbox{E}\kern-.125emX}}
\newcommand*\samethanks[1][\value{footnote}]{\footnotemark[#1]}

\usepackage{algorithm} 
\usepackage{algorithmic}  
\usepackage[algo2e, ruled,linesnumbered]{algorithm2e} 

\setcitestyle{square,numbers}

\begin{document}

\newcommand\relatedversion{}
\renewcommand\relatedversion{\thanks{The full version of the paper can be accessed at \protect\url{https://arxiv.org/abs/1902.09310}}} 

\title{Prescribed Fire Modeling using Knowledge-Guided Machine Learning for Land Management}
\author{Somya Sharma Chatterjee\thanks{University of Minnesota.}
\and Kelly Lindsay\samethanks
\and Neel Chatterjee\samethanks
\and Rohan Patil\thanks{University of California San Diego.}
\and Ilkay Altintas De Callafon\samethanks
\and Michael Steinbach \samethanks[1]
\and Daniel Giron\thanks{Colorado State University.}
\and Mai H. Nguyen\samethanks[2]
\and Vipin Kumar\samethanks[1]
}

\date{}

\maketitle


\fancyfoot[R]{\scriptsize{Copyright \textcopyright\ 20XX by SIAM\\
Unauthorized reproduction of this article is prohibited}}

\begin{abstract}

In recent years, the increasing threat of devastating wildfires has underscored the need for effective prescribed fire management. Process-based computer simulations have traditionally been employed to plan prescribed fires for wildfire prevention. However, even simplified process models are too compute-intensive to be used for real-time decision-making. Traditional ML methods used for fire modeling offer computational speedup but struggle with physically inconsistent predictions, biased predictions due to class imbalance, biased estimates for fire spread metrics (e.g., burned area, rate of spread), and limited generalizability in out-of-distribution wind conditions. This paper introduces a novel machine learning (ML) framework that enables rapid emulation of prescribed fires while addressing these concerns. To overcome these challenges, the framework incorporates domain knowledge in the form of physical constraints, a hierarchical modeling structure to capture the interdependence among variables of interest, and also leverages pre-existing source domain data to augment training data and learn the spread of fire more effectively. Notably, improvement in fire metric (e.g., burned area) estimates offered by our framework makes it useful for fire managers, who often rely on these estimates to make decisions about prescribed burn management. Furthermore, our framework exhibits better generalization capabilities than the other ML-based fire modeling methods across diverse wind conditions and ignition patterns.

\end{abstract}


\section{Introduction}


Thousands of wildfires engulf millions of acres in the United States each year alone \cite{abatzoglou2016impact}. These fires threaten wildlife and human lives and destroy personal property. Fire managers (also known as burn bosses) use prescribed fires, the intentional and controlled lighting of fire, to reduce the fuel that feeds extreme fires and help improve forest health by recycling soil nutrients and controlling pests \cite{wiedinmyer2010prescribed, hiers2020prescribed}. Prescribed fires are generally ignited only when specific conditions are met, such as lower wind speed conditions. Otherwise, lighting of prescribed burns is likely to jeopardize the fire crew's safety and cause collateral damage, similar to the prescribed fire that led to wildfire in Santa Fe National Forest in 2022 \cite{news2022overview}. To minimize the risk of wildfires, fire managers use process-based models to forecast the trajectory of prescribed fires \cite{jain2020review}. These process-based models simulate underlying physical processes in a fire system to identify areas that may burn under given weather conditions (such as wind speed and wind direction), fuel density, and pre-decided ignition pattern. Fire crews use models for such simulations to decide if a prescribed fire can be started under expected wind conditions. However, the weather conditions can change quickly (from the predicted scenario), and the crew has to decide if the risk of a wildland fire starting is big enough to call off the burn. Under such rapidly changing situations, even the fastest fire models cannot enable fire managers to make such decisions.

Among the traditionally used fire models, QUIC-Fire is the only process model specifically designed for prescribed fire simulation and provides the fastest run time at higher resolution \cite{linn2020quic}. QUIC-Fire has been used collaboratively with prescribed fire managers to plan prescribed burns and to understand better the impact of differing weather and ignition conditions on the outcome of a burn. While QUIC-Fire (QF) provides significant speedup over other process-based models \cite{gallagher2021reconstruction}, its computation time and resource requirements still limit its usability for real-time decision-making, which may require rapid assessments of prescribed fire spread as the weather conditions change in real-time. Therefore, a prescribed fire modeling framework with faster than real-time computation time will be useful to fire managers to ensure the safe lighting of prescribed fires.

\begin{figure}[h]
    \centering
    \includegraphics[width=0.47\textwidth]{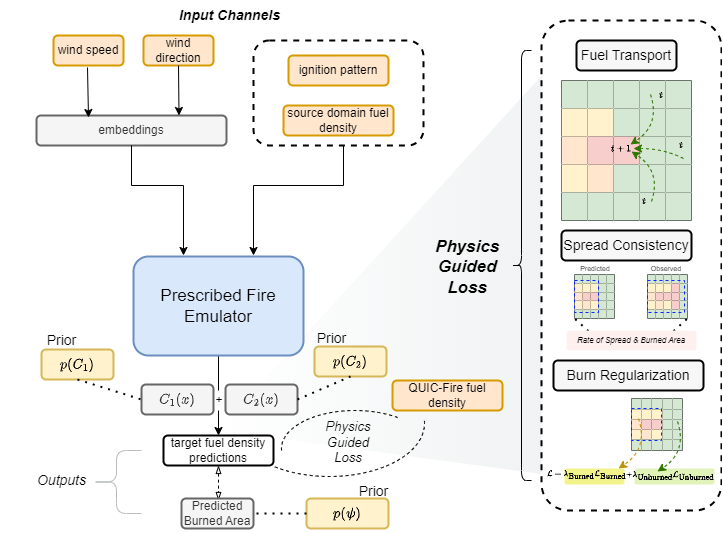}
    \caption{Physics-guided Emulator for fuel density estimation. Inputs include source fuel density, ignition pattern, initial wind speed, and initial wind direction.}
    \label{fig:model-diagram}
    \vspace{-10pt}
\end{figure}

The goal of this paper is to build an ML-based surrogate model that can provide such decisions quickly. While there has been some work on wildfire modeling (see related work), there is little work that can be used in the context of prescribed fire modeling. While a framework using CNN \cite{cope2021using} has been developed for forecasting prescribed fire spread, its utility is limited since it requires simulated variables from QUIC-Fire as input to make forecasts further. Therefore, there is still a need for a framework that can be used to forecast the spread of prescribed fire from the initial wind condition at the time of lighting of fire and still be fast enough to be useful in an operational setting. Additionally, in the context of wildfires, CNN \cite{cope2021using, akagic2022lw, yang2021predicting, marjani2023large, sung20203d}, ConvLSTM \cite{burge2020convolutional, huot2020deep} and UNet \cite{huot2020deep, zhang2021learning} models have been developed to model the evolution of wildfires. However, due to several differences in the exogenous variables, ignition patterns, and fire spread mechanisms, methods developed for wildfires are not suitable for modeling prescribed fires. Wildfires and prescribed fires differ in their physical behavior and environmental conditions, with wildfires being long-lasting fires ignited under low moisture and higher wind speeds at one or few ignition points, while prescribed fires are intentionally set by burn bosses under controlled conditions, often involving multiple fire lines interacting with each other. 

Moreover, the aforementioned ML methods are unable to accurately model the complex nonlinear interaction among physical processes in a prescribed fire system. These methods face three major challenges. First, this leads to physical inconsistencies in the predicted fuel density distribution, such as inaccurate fuel transportation and inconsistent contraction of burned area. Another challenge is forecasting the sequence information from only the initial environmental conditions. In prescribed fire modeling, predictions from ML models suffer from class imbalance problem where the majority class of unburned pixels bias the fuel predictions, translating into an overestimation of fuel in burned regions. This bias problem is exacerbated when testing model performance in more extreme wind conditions in out-of-distribution settings. The third challenge is the inaccurate estimation of summary statistics used by fire managers in operational settings. One such metric is the burned area or number of burned cells that enables an improved understanding of the severity of the burn. However, when the ML model inaccurately predicts the fuel density spatial distribution, the number of burned cells is often underestimated.



To overcome the above challenges, we propose using a physics-guided spatiotemporal model that addresses the complexities of emulating prescribed fire evolution. It integrates physical constraints, leverages domain knowledge, and provides fast time-series forecasts, making it a valuable tool for burn bosses and stakeholders involved in prescribed fire planning and decision-making. In our framework, we propose integrating known physical constraints and domain knowledge in the model to enable the ML model to learn physically consistent values \cite{willard2022integrating}. Such a \textbf{physics-informed surrogate model} would be extremely useful for planning fires and providing speedup over simulators in complex scenarios under different wind conditions, long time scales, or large land areas. To overcome the second challenge, we leverage preexisting source domain data to accurately learn the spread of fire beyond the ignited cells. This \textbf{domain adaptation} mechanism allows us to transfer knowledge for a given ignition pattern from lower wind speed settings to scenarios with a higher wind speed. This enables us to generalize the predictions beyond the case where we can use historical data on the same ignition pattern lit under different weather conditions. To address the third challenge, we integrate a probabilistic graphical structure to model the interdependence between fuel density and the number of burned cells. This ensures that we can enforce prior beliefs about these quantities and that the learned embeddings can capture the hierarchical relationship between the two quantities. The \textbf{probabilistic graphical modeling} and the physics-guided learning ensure that the summary statistics, like the number of burned cells in the grid are estimated accurately. These statistics are used by the fire crew to understand how to modify the ignition pattern to achieve the desired results from the prescribed fire.

We use simulations from QUIC-Fire (perceptual model for fuel cell illustrated in Appendix, Figure~\ref{fig:infographic}), as the ground truth for learning an ML emulator. The proposed ML-based emulator provides speedup over QUIC-Fire and can be used for \textit{faster than real-time prescribed fire forecasting}. From initial wind conditions and user-provided ignition pattern (sequence map of how the burn bosses will ignite cells in a spatial grid), the emulator can provide the evolution of fire (in terms of fuel density, the mass of fire fuel per unit volume) over time. We have released our code in a temporary Google drive \href{https://drive.google.com/drive/folders/1kRRH_7an68mQGHUOcbilwupTdn5wyjhX?usp=sharing}{link}.


The summary of contributions of this work is provided next.

\begin{itemize}
    \item To the best of our knowledge, we propose the first prescribed fire emulator to predict fuel density evolution from initial wind conditions and ignition patterns. The physics-guided model improves the MSE in fuel density predictions by 55\% over data-driven ML models and reduces the fuel density estimation time by 69\% over QUIC-Fire. 
    \item We enrich the feature set by including preexisting source domain data (fuel density similar in ignition pattern) as an additional channel. This ensures that the model overcomes the bias towards the overestimation of fuel and learns a more accurate fire spread behavior. 
    \item Using physics-guided learning, we incorporate physical constraints in the learning process to improve our generalization ability using limited input data while ensuring physical consistency in our predictions. This physical knowledge is drawn from the assumptions and fire behavior encoded in the QUIC-Fire model.
    \item We also estimate fuel density and burned area as part of a graphical modeling structure that improves the representations learned for these quantities at different levels of abstraction.
    \item Extensive experiments demonstrate the effectiveness of the proposed method in emulating prescribed fire behavior. Measuring success in many scientific domains, including fire modeling, is hard to capture via standard performance metrics commonly used in ML model design. We propose new loss functions and evaluation metrics to evaluate the physical consistency of the outputs from ML models. The results of these experiments demonstrate that our approach performs better than other fire modeling methods in terms of both prediction performance and physical consistency of the outputs.
\end{itemize}

\section{Related Work}

Traditional fire behavior simulator models \cite{finney1998farsite,linn2002studying, tymstra2010development, tolhurst2008phoenix, gaudreau2016borealfiresim}  have been used for simulating wildfires \cite{johnston2008efficient,sullivan2009wildland,gollner2015towards,katan2021abwise}. While these models have been adapted for generating prescribed fire simulations, they often make assumptions that may not be suitable for prescribed fire modeling and overlook the fire suppression behavior specific to prescribed fires. Additionally, their computational cost hinders their practical implementation for real-time operational decision-making \cite{schumaker2022hexfire, frangieh2018numerical}. QUIC-Fire \cite{linn2020quic} overcomes some of these challenges and offers faster runtime compared to other process-based models.

Machine learning (ML) models have emerged as computationally efficient alternatives to process-based models, demonstrating success in various wildfire modeling tasks, including fuel characterization, fire risk, and fire effect prediction \cite{jain2020review, zhang2022dynamic, zan2022emulation, bolt2022spatio}, long-term planning \cite{bao2015optimizing, mcgregor2016fast, penman2011bayes}, burned area prediction \cite{cortez2007data, safi2013prediction, storer2016pso, liang2019neural}, fire behavior prediction (e.g., fire spread, growth, severity) \cite{subramanian2017learning, chetehouna2015predicting,kozik2013adaptive}. However, methods developed for wildfire modeling may not generalize to prescribed fire modeling \cite{hiers2020prescribed, furman2018firetec, hoffman2018advancing, cope2021using,linn2020quic}. Wildfires are unplanned and may be caused by natural or accidental human ignitions. Prescribed fires are planned to meet management objectives and are ignited under pre-determined environmental conditions. In terms of fire mechanics, wildfires are often modeled as a single ignition point, eventually leading to an ellipsoid fire shape. In contrast, prescribed fires are ignited at several ignition points, often in terms of several fire lines. This creates a more complex interaction among the fire ignited at different points, leading to fire suppression. Prescribed fires are more complex and difficult to model compared to wildfires. Wildfire modeling often also looks at wildfires that have been burning for days, while prescribed fires are planned to be executed in a single day. Due to differences in time scales, wildfires are more heavily impacted by moisture content in the atmosphere and weather patterns over several days, while the effect of these exogenous variables may not be that prominent for prescribed fires. This leads to a difference in fire behavior in a wildfire and a prescribed fire \cite{hiers2020prescribed, hoffman2018advancing,furman2018firetec}. 

Advancements in physics-guided ML have also contributed to wildfire modeling \cite{allaire2021emulation, bottero2020physics}. However, the physical assumptions made by current methods are not suitable for prescribed fire modeling. These methods overlook the interaction of fire with ambient wind conditions and may suffer from numerical instability \cite{bottero2020physics}. In contrast, our framework leverages simulations from QUIC-Fire as ground truth, enabling the learning of the two-way interaction between environmental conditions and fire behavior without making strong assumptions. Moreover, previous studies in prescribed fire modeling have not extensively explored the specific problem of prescribed fire evolution modeling based on initial environmental conditions \cite{cope2021using, agastra2020evaluating, afrin2021machine, perez2020evaluation}. This capability is crucial for effective prescribed fire planning. By addressing this gap, our framework fills an important need in the existing ML literature. Relevant ML methods for prescribed fire and wildfire forecasting include CNN \cite{cope2021using, akagic2022lw, yang2021predicting, marjani2023large, sung20203d}, ConvLSTM \cite{burge2020convolutional, huot2020deep}, and UNet \cite{huot2020deep, zhang2021learning} models. We also show a comparison with these methods in the paper.

\section{Problem Setting}



In this work, we study the evolution of fuel density, the mass of fire fuel per unit volume, as prescribed fire spreads. We use $N$ simulation runs from the QUIC-Fire model as training examples to learn a data-driven emulator of QUIC-Fire. The input tensor for training example, $i$, can be represented as $\mathcal{X}_i \in \mathbb{R}^{n \times M \times P \times C}$, where $n$ is the number of time steps in the sequence, $M$ and $P$ are the number of rows and columns of cells forming the spatial grid and $C$ is the number of input channels. In our experiments, $C$ includes initial wind speed, initial wind direction, ignition pattern, and source domain fuel density maps. Similarly, the fuel density output for the model can be represented as $\mathcal{Y}_i \in \mathbb{R}^{ n \times M \times P \times 1}$. The initial wind speed and wind direction are static channels - available only for the first time step and repeated throughout the grid and the sequence to obtain the embeddings that are used as inputs to the emulator. Whereas ignition patterns and source domain maps are dynamic inputs.

\section{Methods}
This section details the proposed physics-informed framework (Fig.~\ref{fig:model-diagram}) for fuel density prediction. Section~\ref{sec:method-spatiotemporal} introduces the spatiotemporal model to learn fuel density evolution. Section~\ref{sec:IK} introduces the physical constraints used to improve the physical consistency of predictions. Section~\ref{sec:DA} outlines the source domain data used for data augmentation. In Section~\ref{sec:method-PGM}, we outline the probabilistic graphical modeling structure for incorporating prior knowledge relating to fuel density estimation and burned area estimation. This work studies the driver-response relation in the prescribed fire system. Each simulation run represents different initial wind conditions and ignition patterns. For each simulation run $i$, $\mathcal{X}_i = [x_i^1, x_i^2,...,x_i^T ]$ represent the spatiotemporal drivers, where $x_i^t \in \mathbb{R}^{M \times P \times C}$. Similarly, the response in a simulation run $i$ at time step $t$ can be represented as $y_i^t \in \mathbb{R}^{M \times P \times 1}$. 

\subsection{Spatiotemporal Model}
\label{sec:method-spatiotemporal}

In this work, convLSTM \cite{shi2015convolutional} models provide a pathway to leverage both spatial correlations among the cells in a spatial grid and temporal relation in time-series measurements for each cell. Our spatiotemporal model uses ConvLSTM layers to encode information in the three-dimensional domain - time, height (rows of cells), and width (columns of cells). The ConvLSTM uses the following set of equations to generate embeddings for a sequence,

\vspace{-10pt}
\begin{equation}
\footnotesize
    \begin{split}
        \boldsymbol{i^t}    &= \sigma (\boldsymbol{W_i}\left[[\boldsymbol{x^t}];\boldsymbol{h^{t-1}}; \boldsymbol{c^{t-1}}\right] + \boldsymbol{b_i}),\\
        \boldsymbol{f^t}    &= \sigma (\boldsymbol{W_f}\left[[\boldsymbol{x^t}];\boldsymbol{h^{t-1}}; \boldsymbol{c^{t-1}}\right] + \boldsymbol{b_f}),\\
        \boldsymbol{g^t}    &= tanh (\boldsymbol{W_g}\left[[\boldsymbol{x^t}];\boldsymbol{h^{t-1}}\right] + \boldsymbol{b_g}),\\
        \boldsymbol{c^t}    &= \boldsymbol{f^t} \odot \boldsymbol{c^{t-1}} + \boldsymbol{i^{t}} \odot \boldsymbol{g^t},\\
        \boldsymbol{o^t}    &= \sigma (\boldsymbol{W_o}\left[[\boldsymbol{x^t}];\boldsymbol{h^{t-1}}; \boldsymbol{c^{t}}\right] + \boldsymbol{b_o}),\\
        \boldsymbol{h^t}    &= \boldsymbol{o^t} \odot \tanh{(\boldsymbol{c^t})}.\\
    \end{split}
\end{equation}

\noindent Here, inputs ${x}^t$, cell states $\mathbf{c}^t$, hidden states ${h}^t$ and gates ${i}^t$, $\mathbf{o}^t$, $\mathbf{f}^t$ are 3D tensors. Similar to standard LSTM cell, ConvLSTM cells contain a cell state $\mathbf{c}^t$ that preserves the memory from the past. The forget gate $\mathbf{f}^t$ filters the information obtained from $\mathbf{c}^{t-1}$, and the input gate filters information from the cell state. The new cell state and hidden state are computed as $\mathbf{c}^t$ and $\mathbf{h}^t$. Predicted fuel density is estimated from the hidden units as $\hat{y} = \mathbf{W}_y \mathbf{h}^t$. Therefore, each hidden state $\mathbf{h}^t$ is obtained from hidden and cell states from the prior time step as $\mathbf{h}^{t-1}$ and $\mathbf{c}^{t-1}$. These ConvLSTM layers are used as building blocks to model complex, non-linear interactions in the data. In our model, we stacked four sets of ConvLSTMs and batch normalization layers with ReLU activation. The output layer is a convolution layer that outputs a 3D tensor with one channel for fuel density. This spatiotemporal structure allows the model to determine the future state of a cell in the spatial grid from the inputs and past states of its local neighbors.

\subsection{Transferring Knowledge from Source Domain} \label{sec:DA}

In a prescribed fire system, the complex interaction of fire lines is challenging to model, especially under extreme wind conditions, which make the dynamics of the fire spread more chaotic. Due to a significant imbalance between the number of cells that fall under the burned region compared to the unburned region, conventional ML models revert to predicting all cells to be unburned, leading to an overestimation of fuel in burned regions. Apart from proposing asymmetric loss formulation (see next section), we propose data augmentation using pre-existing source domain data to address the class imbalance bias. In our framework, the source and target domains are fuel density data that are similar in their ignition patterns used for lighting the prescribed fires. However, the environmental conditions, including wind speed and direction, may differ. To maintain a realistic operational setting, we use the same source maps during inference as well. In our experiments, to study the impact of source data on training, we evaluate the model performance with different source data scenarios: (a) no source data, and (b) source fuel density generated with an initial wind speed of 1 m/s and wind direction 230$^\circ$ (\textit{standard source data setting}, Appendix). While source maps with higher wind speeds and disparate wind conditions were also explored, the standard source maps resulted in the most skillful prediction model. Since higher wind speeds accelerate the spread of fire and cause complex interactions between fire and atmosphere, other source settings increased uncertainty and added bias toward the over-burning of fuel. Moreover, since we use one source setting for training, predicting all the target fuel densities with higher wind speeds and disparate wind directions involves overcoming distributional shifts. In addition, fire also influences wind in the system by releasing energy and increasing the air temperature. While these initial ambient wind conditions affect fire behavior, the influence of fire on wind and its feedback on fuel density evolution is not well known due to limited input channels. To overcome this, we integrate known physical constraints to improve predictions from the model. We discuss this further in the next section.  

\subsection{Integrating Knowledge} \label{sec:IK}

\subsubsection{Transferring knowledge from the physics-based model: The problem of fuel transport}
\label{sec:method-FT}

\begin{definition}[Fuel Transport]
    The problem of fuel transportation relates to the transportation of fuel by the wind from one location to another. For fuel density $y_s$ at location $s$, fuel transportation results in $y_{t, s} = f(y_{t-1, s})+ \rho y_{t, s'}$, where $f$ represents the non-linear effect of other physical processes on fuel density, $\rho$ refers to the fraction of fuel that is transported from location $s'$ to $s$. Thus, the probability of fuel increasing in an ignited cell $s$ is non-zero, $P(y'_{t,s} > y'_{t-1,s})> \delta$ where $\delta$ is an arbitrary, non-negative value and $y'$ refers to fuel density in an ignited cell.
\end{definition}

Fuel transport is more common in the case of wildfires that generally spread under higher wind speeds. Since prescribed fires are lit under very specific wind conditions (low wind speed), fuel transportation is unlikely. This assumption is encoded in the QUIC-Fire model. In simulations generated by QUIC-Fire with a homogenous initial fuel profile, this helps avoid any uncharacteristic expansion and contraction of burned fuel density arising from fuel transportation from observed or unobserved regions. To emulate this behavior, we incorporate the assumption that a cell's fuel density cannot increase over time by fuel transportation. Therefore, we penalize those predictions that have an uncharacteristic increase in fuel density over time, given as, 

\begin{equation} \footnotesize
    \mathcal{L_{FT}} = \frac{\sum_t||Y_t - \hat{Y}_{t}|| \odot \mathds{1}((\hat{Y}_t - \hat{Y}_{t-1})>\epsilon)}{T}.
\end{equation}

\noindent To put a soft constraint, $\epsilon$, a non-negative value, is used to set the tolerance for what will be defined as a significant increase in fuel density over two consecutive time steps. $T$ is the number of time steps

\subsubsection{Spread Consistency} 
\label{sec:method-FM}

Accurately predicting the spread of prescribed fire is essential in ensuring the fire crew can safely control the blaze while meeting the burn objectives. Fire managers often rely on estimates of fire spread metrics like rate of spread and burned area indices to understand how the fire is going to spread. Since these quantities are estimated from fuel density, we regularize those fuel density predictions that lead to higher errors in estimates for the fire spread metrics. We, therefore, ensure the spread consistency by estimating the average rate of spread and burned area percentage from the predicted fuel density and minimizing the difference with the estimates from the QUIC-Fire simulations.


\begin{definition}[Rate of Spread]
    Average rate of spread for each time step is computed as the distance traveled by fire over the time spent from the initial time step to time $t$  \cite{sullivan2020wildland}, $\mathcal{{ROS}}(Y_t) = {\text{Distance}}/{\text{Time Taken}} = {(\eta({Y_t}) - \eta({Y_{t_0}}))}/{(t - t_0)}$. Here, $t_0$ is the initial time step, and $t$ is the time step for which ROS is being computed, $\eta(.)$ estimates the number of columns impacted by the fire.
\end{definition}

Here, $Y_{t_0}$ refers to where the fire is ignited. In the dataset, the wind generally starts from the west and blows toward the east. This allows us to estimate a unilateral rate of spread based on how many columns of cells are impacted by the fire, enabling us to reduce the number of FLOPS for ROS estimation in the loss computation. 

\paragraph{Rate of Spread (ROS) loss} The mean squared error in the rate of the spread between predicted and observed fuel density values is minimized as part of the loss, as 

\begin{equation} \footnotesize
    \mathcal{L_{ROS}}=  \frac{\sum_t (\mathcal{{ROS}}(Y_t) - \mathcal{{ROS}}(\hat{Y}_t))^2}{T}.
\end{equation}

\begin{definition}[Burned Area Percentage]
    Burned area percentage refers to the percentage of cells burned in the whole grid at time $t$, $\mathcal{{BA}}(Y_t) = {(\sum_i^M \sum_j^P \mathds{1}(Y_{i, j, t}< \epsilon_b) \times 100 )}/{(M \times P)}$.
\end{definition}

\paragraph{Burned Area (BA) Percentage loss} Similar to the rate of spread loss term, the mean squared error between the predicted and observed burned area metric is minimized, as

\begin{equation} \footnotesize
    \mathcal{L_{BA}} = \frac{\sum_t (\mathcal{{BA}}(Y_t) - \mathcal{{BA}}(\hat{Y}_t))^2}{T}.
\end{equation}

Incorporating $\mathcal{L_{ROS}}$ and $\mathcal{L_{BA}}$ ensures that the statistical properties of the predicted fuel density match the ones of the QUIC-Fire simulations.

\subsubsection{Burn Regularization} 
\label{sec:method-BR}

Due to limited input data, the fuel density predictions from data-driven models do not generalize well to out-of-distribution (OOD) wind conditions. Standard spatiotemporal models also face the issue of over-burning fuel in unburned cells. We can penalize this behavior in the loss term by including a weighted loss on the unburned cells. Similarly, in the absence of knowledge about the physical behavior of fire and its interaction with fuel moisture, standard ML model predictions exhibit the under-burning of fuel in cells that have started burning. To address this, we can add weighted loss that penalizes under-burning behavior in burned cells. The burned loss is given as follows,

\begin{equation} \footnotesize
    \mathcal{L_{\text{Burned}}} = \frac{\sum_t||Y_t - \hat{Y}_{t}|| \odot \mathds{1}((Y_t )<\epsilon_b)}{T},
\end{equation}

Similarly, to address the model bias towards lower fuel density values under high  wind speeds, we can add weighted loss to regularize the over-burning behavior in unburned area. The unburned loss is given as follows,

\begin{equation} \footnotesize
    \mathcal{L_{\text{Unburned}}} = \frac{\sum_t||Y_t - \hat{Y}_{t}|| \odot \mathds{1}((Y_t )>\epsilon_u)}{T}.
\end{equation}

To enforce these physical constraints, we formulate the physics-guided loss function as,

\begin{equation} \footnotesize
\begin{split}
    \mathcal{L} = & ||Y - \hat{Y}|| + \lambda_{FT} \mathcal{L_{FT}} + \lambda_{ROS} \mathcal{L_{ROS}}  \\
    +  & \lambda_{Burned} \mathcal{L}_{Burned} + \lambda_{Unburned} \mathcal{L}_{Unburned}  \\
\end{split}  
\label{eq:loss0}
\end{equation}

\subsection{Probabilistic Graphical Modeling (PGM)}
\label{sec:method-PGM}

To integrate any prior knowledge about the physical processes, we incorporate a probabilistic graphical modeling structure in our framework. Moreover, this enables us to represent different variables of interest into multiple levels of abstraction, allowing us to leverage the interdependence among these variables for better fuel density estimation. Scientific problems like prescribed fire modeling have a high degree of complexity arising from the interaction of different physical processes. In our case, the complex interaction of ambient wind conditions, fire-generated wind, and fire necessitates the hierarchical structure to improve the modeling of their effect on fuel density.  Here, we model two quantities - fuel density and number of ignited cells. Modeling of the number of ignited cells captures more global similarities in the observed and predicted values, whereas fuel density modeling focuses more on specific or local similarities in the observed and predicted values. 

We incorporate mixture density modeling \cite{bishop1994mixture} to factorize the effect of these multiple physical modalities in the latent space and improve response posterior estimation of fuel density. We model fuel density as a mixture of Gaussian components in our framework. Additionally, incorporating prior knowledge about burned area estimation can also improve fuel density modeling. Drawing inspiration from epidemic modeling \cite{kremer2021quantifying} and wildfire modeling \cite{koh2023spatiotemporal}, where Poisson distribution is used to model count datasets with extreme values, we use Poisson priors to estimate the number of ignited cells in the grid at each time $t$. 


In the following subsections, we describe our probabilistic graphical modeling approach.

\subsubsection{Gaussian Mixture Density Networks}

To model the response as a mixture of Gaussian components, we estimate the parametric distributions $C_j$ and membership of each component represented by the mixing parameter, $\pi_j$, based on the parameter vector learned by the neural network $\phi(x)$. The fuel density response can be represented as,

\begin{equation} \footnotesize
    \hat{y} = \pi_1(\phi(x))C_1(y|\phi(x)) + \pi_2(\phi(x)) C_2(y|\phi(x)),
\end{equation}

\noindent where component $C_j$ can be represented as realizations from a Normal distribution as, $C_j \sim N(\theta(x))$. For each Gaussian component $j$, parameters $\theta = \{\mu, \sigma\}$ and $\pi$ are learned as outputs of the neural network $\phi(x)$. The parameters can be estimated by minimizing the negative logarithm of likelihood \cite{bishop1994mixture},

\begin{equation} \footnotesize
   \mathcal{L}_{MDN} =  - \sum_i  log ( \sum_j (\phi(x_i)) \pi_j(\phi(x_i))C_j(Y_i|\phi(x_i)) )
\end{equation}

\subsubsection{Burned Area Estimation from Poisson Processes}


We also consider the point pattern of fire ignitions as a realization of a Poisson process. For a random count measure $\psi$ that represents the number of cells that are ignited in a grid at time $t$, we model the rate function, $\lambda(y|x,t)$, as $\lambda(y|x,t) = \mathbb{E}(\psi) = \mathbb{E} \sum_i^\psi \mathds{1}(y_i|x_i \in I)$, where $I$ represents the Borel set representing the ignited cells in the observed region at time $t$. Poisson processes characterize the number of ignited cells as estimated from fuel density predictions, $\psi \sim Pois(\lambda(y|x,t))$. We employ the variational free energy function to estimate $\psi$ as,

\begin{equation} \footnotesize
    \mathcal{L}_{PP} = KL(q(\psi)||p(\psi)) - \mathbb{E}_{q(\psi)}[log P(\mathcal{D}| \Phi_{\psi})]
\end{equation}

\begin{figure}[h]
    \centering
    \includegraphics[width=0.3\textwidth]{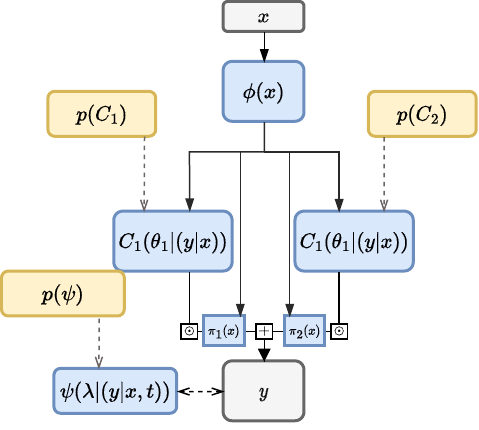}
    \caption{Graphical Model}
    \label{fig:graphmodel}
    \vspace{-10pt}
\end{figure}

The KL term ensures that the posterior learned for $\psi$ is parsimonious to maintain similarity with the prior $p(\psi)$. The negative log-likelihood cost helps ensure that appropriate $\lambda$ is learned to fit $\Phi_{\psi}$ to our dataset $\mathcal{D}$. To incorporate the probabilistic graphical structure in the framework, the loss function can be modified from Eq.~\ref{eq:loss0} to Eq.~\ref{eq:loss}. A perceptual model for the PGM is given in Figure~\ref{fig:graphmodel}.

\begin{equation}
\begin{split}
    \mathcal{L} = & \mathcal{L}_{MDN}  + \lambda_{FT} \mathcal{L_{FT}} + \lambda_{ROS} \mathcal{L_{ROS}}  \\
    + & \lambda_{Burned} \mathcal{L}_{Burned} + \lambda_{Unburned} \mathcal{L}_{Unburned}\\
     + & \mathcal{L}_{PP} \\
\end{split}  
\label{eq:loss}
\end{equation}

\begin{algorithm}
\footnotesize
\caption{\footnotesize Physics-Guided Emulator for Prescribed Fire Simulation}\label{alg:pgcl}
\KwIn{Input Channels $\mathcal{X}_i \in \mathbb{R}^{n \times M \times P \times C} \forall i=1,...,N$ }

\KwOut{Learned neural network estimator $\Phi$ for fuel density prediction, predicted fuel density $\mathcal{\hat{Y}}_i \in \mathbb{R}^{n \times M \times P \times 1}  \forall i=1,...,N$}

\For {batch $\{x_i, y_i\}$ }
{

    estimate $\phi(x_i)$
    
    estimate $\hat{y}_i = \sum_j^k \pi_j(\phi(x_i)) \odot C_j(y_i|\phi(x_i))$

    sample $\psi(\hat{y}_i) \sim Pois(\lambda(y_i|x_i,t))$
    
    Compute the loss as Eq \ref{eq:loss}
    
    Compute gradients and update parameters in $\Phi$ which include $\lambda, \pi_j, C_j$ and $\phi$     
}
\label{alg:pgclplus}
\end{algorithm}

The framework is summarized in Appendix. In each training step, we first estimate the latent representations learned from the spatiotemporal model $\phi(x)$ for each batch. These representations are further used to estimate the Gaussian mixture model components to compute the fuel density $\hat{y}_i$. $\psi$ is sampled from $Pois(\lambda)$ where $\lambda$ depends on the learned conditional response distribution $y|x$ at each time $t$. To update all the parameters in the model $\Phi$, compute gradient with respect to the loss function given in Eq.~\ref{eq:loss} and update all the parameters. The model without the graphical model component is called PGCL (loss Eq.~\ref{eq:loss0}), while the one with the graphical modeling component is called PGCL+ (loss Eq.~\ref{eq:loss}).

\section{Results}
\subsection{Experimental Setup}

\subsubsection{Baselines}

We compare the proposed models in the methods section with other baselines in wildfire and prescribed fire modeling. We compare our results with CNN \cite{cope2021using, akagic2022lw, yang2021predicting, marjani2023large, sung20203d}, ConvLSTM (CL) \cite{burge2020convolutional, huot2020deep} and UNet \cite{huot2020deep, zhang2021learning}. We also include Firefront UNet model \cite{bolt2022spatio} that learns representation for fire state, spatial forcing and weather in separate encoders. We include a CL-GL model that further modifies the CL model with Gram loss-based regularization to match the statistical similarities between the predicted and observed fuel density. Additionally, we include the \textit{Match baselines} that uses historical fuel maps as a prediction on how fuel may change in the future. To estimate fuel in the test set, we consider two baselines - Match ignition baseline and Match wind baseline. Match ignition baseline looks at the first fuel map in the historical data with the closest ignition pattern - while the wind conditions may vary. Match wind looks at the first fuel map in the historical data with the closest wind conditions - while the ignition pattern may vary.

\subsubsection{Dataset and Experimental Details} We use simulation runs from the QUIC-Fire model \footnote[1]{Simulation runs can be generated using the code provided in \href{https://github.com/QUIC-Fire-TT/ttrs_quicfire/tree/main}{this GitHub repository}} to learn the prescribed fire emulator. To test generalization under different environmental factors, the simulation runs include 5 different ignition patterns, 7 wind speeds, and 11 wind directions. We simulate and use 100 runs as training examples. The simulation runs are for a grassland setting with two-dimensional evolution of fires captured in 300 x 300 cells grid over $n$ time steps at 1 second time intervals. Each cell is at a 2m x 2m resolution. In the experimental setup, we randomly split the 100 runs and put 50\% of the data into training and the rest into test dataset, with each comprising 50 samples. Therefore, each of the datasets has input data with dimensionality $50 \text{ simulation runs } \times 50 \text{ time steps }\times 300 \text{ rows } \times 300 \text{ columns } \times 4 \text{ features }$. Wind data is standardized using min-max scaling, whereas fuel density data is not scaled since it varies between 0 and 0.7. With batch size 1 and using the Adam optimization method for gradient estimation, we train each model for 250 epochs \footnote[1]{Code link provided \href{https://drive.google.com/drive/folders/1kRRH_7an68mQGHUOcbilwupTdn5wyjhX?usp=sharing}{here}}. The code is implemented using Tensorflow 2.0 and NVIDIA A40 GPU. Hyperparameters, including penalty coefficients in the loss terms, are fine-tuned using random grid search in the models. Learning rate is 0.001, $\lambda_{FT}, \lambda_{Burned} = 0.001$, $\lambda_{Unburned}, \lambda_{FM} = 0.0001$. In the experiments, we use $\epsilon=$ 0.001 for the physical constraint loss masking. We also use $\epsilon_b=$0.1 and $\epsilon_u=$0.65 for the burned and unburned loss masking, respectively. In the generalization results, we sample test runs into different datasets with different physical properties. This includes sampling based on wind speed, wind direction, and ignition patterns. $\mathcal{D}_{\text{Low Wind}}$ dataset has samples with initial wind speed less than 10 m/s. $\mathcal{D}_{\text{High Wind}}$ dataset has samples with initial wind speed greater than 10 m/s. $\mathcal{D}_{\text{NW Wind}}$ dataset has samples with initial wind direction blowing from the northwest, and $\mathcal{D}_{\text{SW Wind}}$ dataset has samples with initial wind direction originating from the southwest direction. $\mathcal{D}_{\text{Aerial}}$, $\mathcal{D}_{\text{Outward}}$, $\mathcal{D}_{\text{Strip South}}$, $\mathcal{D}_{\text{Inward}}$ and $\mathcal{D}_{\text{Strip North}}$ datasets include samples with different ignition patterns for igniting the fire. 

\subsubsection{Metrics} We evaluate several metrics that compute how well the model predicts the evolution of fire over time. In the tables, the downward arrow indicated that lower values of evaluation metrics indicate better model performance. We evaluate MSE, burned area MSE, unburned area MSE and fire metric MSE (ROS MSE + BA MSE) on test set. We formulate a metric, \textbf{Dynamic MSE (DMSE)}, that evaluates model performance based on change in fuel density over time. For time steps with bigger change in the observed fuel density, we ensure that the error in predictions are penalized more, $\footnotesize DMSE = {(\sum_N (Y_t - Y_{t-1}) \cdot (||\hat{Y}_t - Y_t||)
    )}/{( \sum_N (Y_t - Y_{t-1}))}$. 
We can further validate the physical consistency of the predictions by evaluating how often the physical constraints are met in the predicted values using the following metrics.
\noindent \textbf{Metric}$_{\text{FT}}$ is the percentage of cells that do not follow the fuel transport constraint in the predicted values. \textbf{Metric}$_{\text{Burned}}$ is the percentage of unburned cells that are predicted to be burned. \textbf{Metric}$_{\text{Unburned}}$ is the percentage of unburned cells where fuel densities are underestimated. \textbf{Metric}$_{\text{False Positive}}$ is the percentage of burning cells that are predicted to be burned. \textbf{Metric}$_{\text{False Negative}}$ is the percentage of burning cells that are predicted to be unburned.




\begin{figure*}
\centering

\includegraphics[scale=0.5]{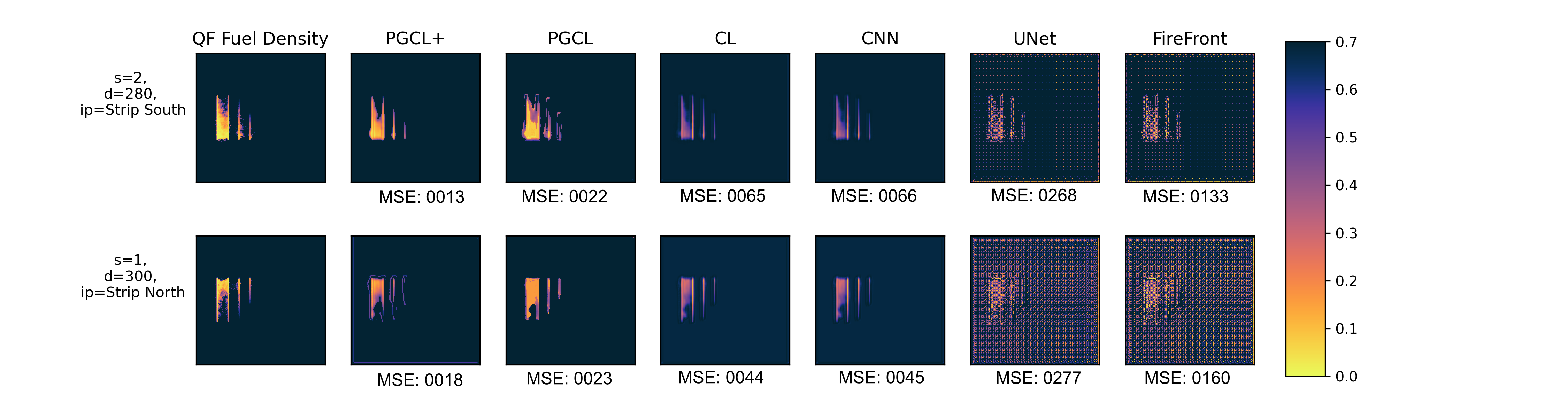}
\caption{Predicted fuel density by different models. s: initial wind speed, d: initial wind direction, ip: ignition pattern. QUIC-Fire (QF) Fuel density is the ground truth we are emulating.}
\label{fig:individual_predictions}
\vspace{-12pt}
\end{figure*}

\subsection{Results and Discussion}

\subsubsection{Source Input Data}

Source fuel density maps enable the model to improve the learning of spatiotemporal changes in target fuel density. Table~\ref{tab:baselines} compares CNN, UNet, CL, CL-GL, and Firefront models in learning fire behavior in the absence and presence of source fuel density data. The MSE values reflect that by including the source fuel density sequence data as one of the input channels, all the models overcome the bias towards over-estimation of fuel. Moreover, the CL model performs better than the other models in predicting fuel density distribution, suggesting that including temporal information and the parsimonious architecture of the model is sufficient for the skillful prediction of fuel density. For further experiments, we explore CL model for physics-guided learning.

\begin{table}[h]
    \centering
    \resizebox{\columnwidth}{!}{
    \begin{tabular}{p{2cm}lllllll}
        \hline
      
        Method & Data & MSE ($\downarrow$) & Burned & Unburned  & Fire Metrics  & DMSE($\downarrow$) \\
         & & & MSE($\downarrow$)& MSE($\downarrow$)& MSE($\downarrow$)&\\\cmidrule(lr){1-7}
        CNN &(w/o source maps)     & 4.9809  & 4.9808  & 9.9618 & 4.9809 & 0.0003  \\
        U-Net & (w/o source maps)   &  2.9754   & 2.9754 & 5.9508& 3.1712 & 0.0002  \\
        CL & (w/o source maps)  &  1.1745   & 1.1745 & 2.3490& 1.1745 & 0.0003  \\
        FireFront& (w/o source maps) & 1.6238    & 1.6238   & 3.2476   & 2.6008  & 0.0013   \\
        CL-GL& (w/o source maps) & 1.7971  & 1.7971  & 3.5943    & 1.7982    & 0.0004  \\
        CNN  & (w source maps)  & 0.0294   & 0.0294  & 0.0594 & 0.0748 & 0.0001  \\
        U-Net & (w source maps)    & 0.0352    & 0.0352  & 0.0704 & 0.0354 & 0.0002  \\
        CL & (w source maps) & \bf{0.0282}   & \bf{0.0282} & \bf{0.0416}   & \bf{0.0209}   & 0.0001 \\
        FireFront& (w source maps) & 0.0293   & 0.0293 & 0.0586 & 0.0745 & 0.0001  \\
        CL-GL& (w source maps) &  0.2336  & 0.2336 & 0.4672   & 0.2342   &0.0002   \\

        \hline
    \end{tabular}
    }
    \caption{\small Test set model performance. CL with source domain input data outperforms other methods.}
    \label{tab:baselines}
\end{table}

\subsubsection{Integrating Physical Knowledge}

\begin{table}
    \centering
    \scriptsize
    \begin{tabular}{p{1.2cm}llllll}
        \hline

        Method & MSE ($\downarrow$)& Burned & Unburned & Fire Metrics & DMSE ($\downarrow$)\\
        && MSE ($\downarrow$)& MSE ($\downarrow$) & MSE ($\downarrow$) & \\
        \cmidrule(lr){1-6}

        CL   &  0.0282  & 0.0282 & 0.0416   & 0.0209   & 0.0001   \\
        PGCL & 0.0157   & 0.0157 & 0.0312 & 0.0307 & 0.0002  \\ 
        PGCL+ & \bf{0.0126}   & \bf{0.0126} & \bf{0.0252}   & \bf{0.0127}   & \bf{0.0001}   \\
        Match Ignition &  0.0275   & 0.0275 & 0.0265  & 0.0279   & 0.0051   \\
        Match Wind & 0.0490   & 0.0490 & 0.0403  & 0.0352   & 0.0048    \\
        
        \hline
       
    \end{tabular}
    \caption{\small Test set model performance. Physics-guided ConvLSTM model (PGCL) and Physics-guided ConvLSTM model + PGM (PGCL+) outperform other methods.}
    \label{tab:losses}
\end{table}

\begin{table}[ht]
    \centering
    \footnotesize
    \resizebox{0.49\textwidth}{!}{
    \begin{tabular}{cccccc}
        \toprule
        \cmidrule(lr){2-6}
         Sample & FT ($\downarrow$) & Burned ($\downarrow$)& Unburned ($\downarrow$)& \begin{tabular}{@{}c@{}}False \\ Positive ($\downarrow$)\end{tabular} & \begin{tabular}{@{}c@{}}False \\ Negative ($\downarrow$)\end{tabular} \\
         \midrule

         CL  & 0.07 & 20.30 & 57.51 & 20.21 & 50.03\\
         PGCL  & 0.03 & 2.59 & 1.79 & 2.63 & 38.55\\
         PGCL+ &  \bf{0.01} & \bf{0.12} & \bf{0.44} & \bf{0.18} & \bf{33.02}\\
         \bottomrule
    \end{tabular}
    }
    \caption{\small Physical consistency evaluations metric for test samples. s=wind speed, d= wind direction. }
    \label{tab:pred_label}
\end{table}

\begin{table}[h]
    \centering
    \scriptsize
    
    \begin{tabular}{p{1cm}lllllll}
        \hline

         & QUIC-Fire &  PGCL+ & PGCL & CL & CNN & UNET \\ \cmidrule(lr){1-7}
        Minimum & 28 s & 8 s & \bf{6} s & 24 s & 12 s & 16 s \\
        Mean & 42 s & 13 s & \bf{12} s & 16 s & 13 s & 17 s \\
        Maximum & 116 s & 20 s & 17 s & 27 s & \bf{16} s & 19 s \\
        \hline

    \end{tabular}
    
    \caption{\small Model inference time summary statistics for predicting test sequence with 50 time steps (in seconds) averaged over 10 repetitions and all test samples}
    \label{tab:inference-time}
    \vspace{-10pt}
\end{table}

\begin{figure*}[ht]
\centering
\begin{tabular}{cccc}
\centering
 \subcaptionbox{}{\includegraphics[scale=0.22]{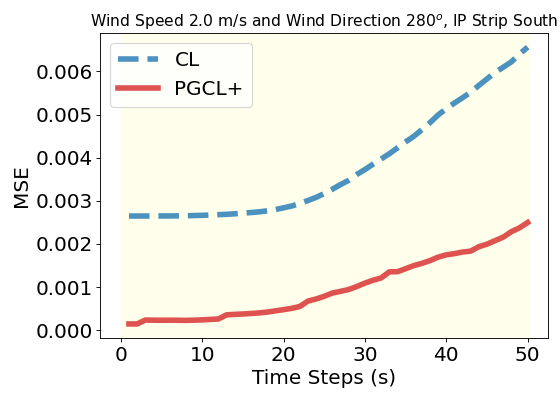}} &

 \subcaptionbox{}{\includegraphics[scale=0.22]{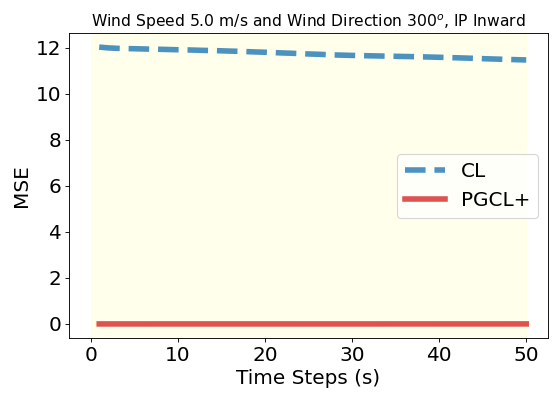}} &

\subcaptionbox{}{\includegraphics[scale=0.22]{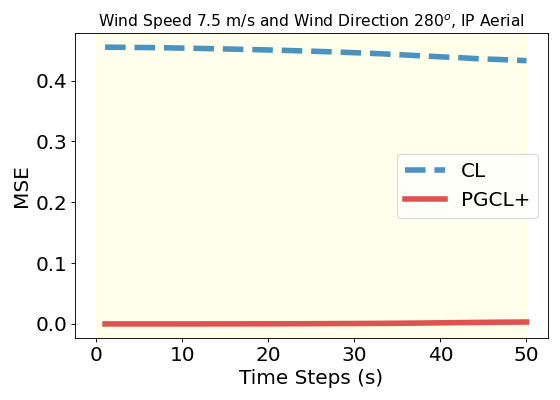}} &
 
 \subcaptionbox{}{\includegraphics[scale=0.22]{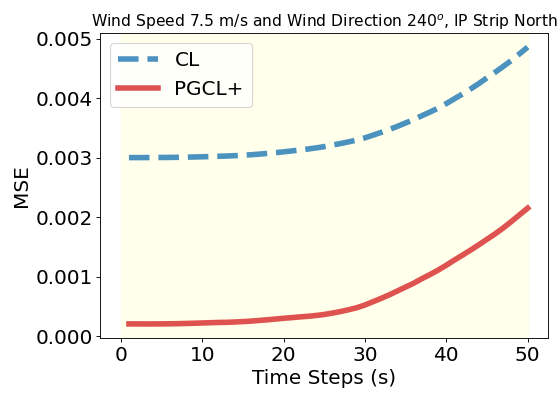}} \\


 

\end{tabular}
\caption{\small \textbf{Individual Predictions}: Generalization over time in test samples  }
\label{fig:msebytime-comparison}
\vspace{-10pt}
\end{figure*}

We investigate the effect of incorporating physical constraints on prescribed fire emulation. The model performance in terms of the test set evaluation metrics is given in Table~\ref{tab:losses}. The physics-guided models outperform other baseline models. PGCL and PGCL+ also show a reduction in fire metric MSE values suggesting that the overall rate of spread and burned area percentage in predictions is similar to the observed values. 

While the MSE values are useful for evaluating the average model performance, investigating individual prediction examples can be useful to ensure generalization under different distributional settings. Figure~\ref{fig:individual_predictions} shows the individual predictions at the 50$^{\text{th}}$ second and also outlines the test sample attributes for comparison. We compare the PGCL+, PGCL, CL, CNN, UNet, and FireFront models. We notice that the physics-guided model has the most stable predictions while other methods are over-burning unburned cells and under-burning burned areas. In models that are not informed by physical constraints in the loss term, we also notice higher fuel density predictions in cells surrounding the fire perimeter. This behavior arises due to complex wind dynamics in that region that is not being captured by these models. Without physics guidance, the standard ML models inaccurately predict that region's fuel density. Furthermore, in the cells predicted to be burned, the physics-guided model accurately predicts the lower fuel density values. The physics-guided model achieves the most accurate predictions for fuel density values among the models.

We can further compare the model performance in their ability to capture the physical behavior of the fire accurately. Results in Table~\ref{tab:pred_label} summarize physical consistency metrics for test samples. The results suggest that the physics-guided models (PGCL, PGCL+) achieve better physical consistency in the predictions. The physics-guided models have a lower proportion of cells where the fuel transport constraint is not met. The physics-guided models also have a lower proportion of falsely predicted burned cells and falsely predicted unburned cells. This is also evident in the visualizations of the individual predictions. The physics-guided model provides better burn predictions than the standard ConvLSTM model, as shown in the lower values for the physical consistency evaluation metrics in Table~\ref{tab:pred_label}. 

In Table~\ref{tab:inference-time}, we also compare the proposed framework with QUIC-Fire in terms of inference time for predicting 50 seconds of fuel density. We see that PGCL+ achieves 3-5 times speedup compared to QUIC-Fire.  

\begin{figure}
\centering

\includegraphics[scale=0.19]{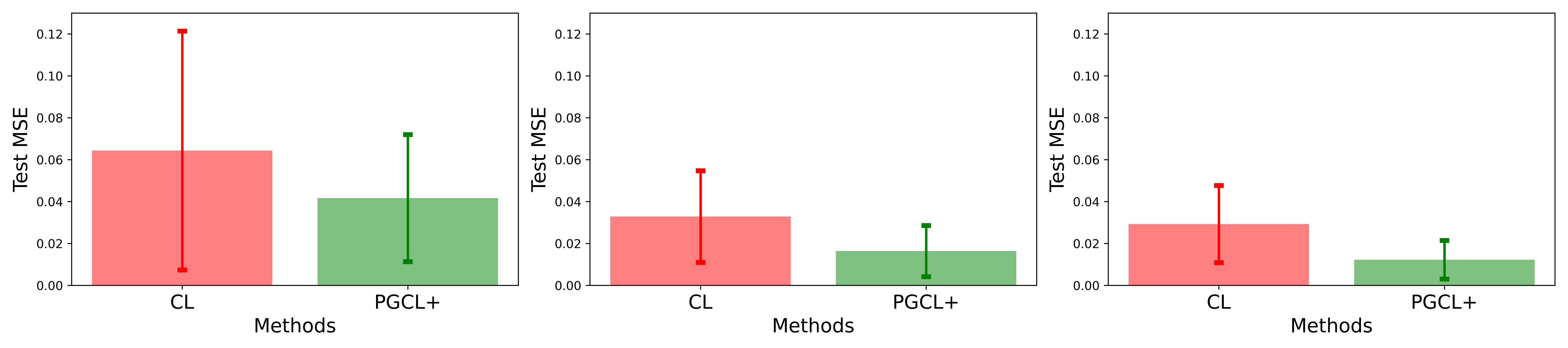}

\caption{Model performance under different number of examples from the train set used for training. From left to right, subplots for 20\% training data, 50\% training data and 100\% training data}
\vspace{-15pt}
\label{fig:datascarcity}
\end{figure}

\subsection{Generalization under Different Environmental Conditions}

\begin{figure}
\centering

\includegraphics[scale=0.28]{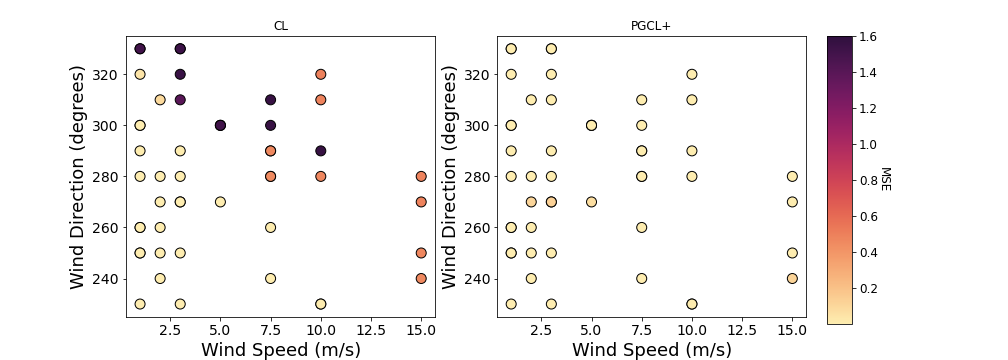}
\caption{Average MSE by wind conditions for CL and PGCL+. Color represents the MSE values.}
\label{fig:msebywind}
\vspace{-15pt}
\end{figure}

Figure~\ref{fig:msebywind} summarizes test set MSE values by the initial wind direction and wind speed conditions. The physics-guided model outperforms the data-driven spatiotemporal model. Especially under higher wind speed conditions and under wind direction greater than 280$^\circ$, the CL model has higher test MSE values. This is because higher wind speeds introduce increased uncertainty and complexity into fire behavior prediction. This is further evident in Table~\ref{tab:losses-generalization}, where we report the test MSE values by different wind speeds, wind directions, and ignition patterns to further evaluate the generalization under different environmental conditions. In general, PGCL+ achieves better model performance. Integrating physics has a bigger impact on the samples, wherein the initial wind direction originates from SW. Model performance for test samples with wind originating from NW is more difficult due to the distributional shift between the target and the source maps since source maps include winds originating from SW. 

\begin{table}[h]
    \centering
    \scriptsize
    
    \begin{tabular}{p{2cm}lllll}
        \hline

         Category & Data &  PGCL+  & PGCL & CL \\ \cmidrule(lr){1-5}
          \multirow{2}{*}{\bfseries Wind Speed} & $\mathcal{D}_{\text{Low Speed}}$ & \bf{0.0073} &  0.0115& 0.0103  \\
           & $\mathcal{D}_{\text{High Speed}}$ & \bf{0.0177} & 0.0317  & 0.1142  \\
           \hline
          \multirow{2}{*}{\bfseries Wind Direction} & $\mathcal{D}_{\text{NW Wind}}$ & \bf{0.0290} & 0.0293 & 0.0302 \\
           & $\mathcal{D}_{\text{SW Wind}}$ & 0.0042  & \bf{0.0010} & 0.0172 \\
           \hline
          \multirow{5}{*}{\bfseries Ignition Pattern} & $\mathcal{D}_{\text{Aerial}}$ & \bf{0.0054} & 0.0194 & 0.0411\\
           & $\mathcal{D}_{\text{Outward}}$ & \bf{0.0028} &  0.0065  & 0.0419 \\
           & $\mathcal{D}_{\text{Strip South}}$ &\bf{0.0014} & 0.0608  & 0.0180 \\
           & $\mathcal{D}_{\text{Inward}}$ &0.0301 &  \bf{0.0091}  & 0.0122 \\
           & $\mathcal{D}_{\text{Strip North}}$ &\bf{0.0022}  & 0.0637 & 0.0077\\

        \hline
    \end{tabular}
    
    \caption{\small Generalization under different physical conditions.}
    \label{tab:losses-generalization}
    \vspace{-8pt}
\end{table}

Figure~\ref{fig:msebytime-comparison} further evaluates the test MSE at each time step predicted in the fuel density time series. We see for some samples, the MSE increases over time due to increased variability in the fire spread. We also notice that the physics-guided model achieves lower MSE than the data-driven model at all time steps.

In the real-world setting, the availability of training data may also impact model performance. We investigate the model performance when different quantities of examples are available from the training dataset. Figure~\ref{fig:datascarcity} shows the test set MSE values when the proportion of train set samples available for training varies. We notice as the data scarcity reduces, the model performance improves. The physics-guided model is also able to achieve a lower test set MSE than the CL model.

\section{Conclusion}
This work proposes a novel knowledge-guided spatiotemporal model that integrates physical knowledge and uses source fuel maps to emulate the fuel density changes under different wind conditions and ignition patterns. Our approach, which uses a loss function with physical constraints, reduces or eliminates different physical inconsistencies in predictions, including fuel transport, over-burning in unburned areas, and over-estimating fuel in burned areas. Our framework can be extended to other spatiotemporal models for learning physically consistent fuel densities. Since the framework achieves faster-than-real-time inference, it enables avenues for fire managers to adapt their plans to achieve a burn's intended objective with changing environmental conditions.

Several future directions can be explored. The proposed method can be modified to include other physical laws and physical constraints. For instance, the work can be extended to three-dimensional grids that include height as a dimension. Here we can also include canopy consumption to track how the fuel density evolves across the three-dimensional spatial grid. The framework can also be trained on longer time series to ensure that burn bosses can obtain reliable prescribed burn predictions for extended time periods.

\section{Acknowledgments}
    This work was funded by the NSF awards 2134904 and 1934721. Access to computing facilities was provided by the Minnesota Supercomputing Institute.


\bibliographystyle{ACM-Reference-Format}
\begin{footnotesize}
\bibliography{main-bib}
\end{footnotesize}




\end{document}